\documentclass{article}
\usepackage{spconf,amsmath,graphicx,bm}
\usepackage{hyperref}
\usepackage{multirow}
\usepackage{booktabs}
\usepackage{algorithm,algorithmic}
\hypersetup{hypertex=true,
	colorlinks=true,
	linkcolor=black,
	anchorcolor=black,
	citecolor=black}

\title{Prototype-Based Layered Federated Cross-Modal Hashing}
\name{Jiale Liu$^{1}$, Yu-Wei Zhan$^{1}$, Xin Luo$^{1}$, Zhen-Duo Chen$^{1}$, Yongxin Wang$^{2}$, Xin-Shun Xu$^{1}$}
\address{$^{1}$School of Software, Shandong University  \\ $^{2}$ School of Computer Science, Shandong Jianzhu University}
\begin{document}
%
\maketitle
\begin{abstract}
Recently, deep cross-modal hashing has gained increasing attention. However, in many practical cases, data are distributed and cannot be collected due to privacy concerns, which greatly reduces the cross-modal hashing performance on each client. And due to the problems of statistical heterogeneity, model heterogeneity, and forcing each client to accept the same parameters, applying federated learning to cross-modal hash learning becomes very tricky. In this paper, we propose a novel method called prototype-based layered federated cross-modal  hashing. Specifically, the prototype is introduced to learn the similarity between instances and classes on server, reducing the impact of statistical heterogeneity (non-IID) on different clients. And we monitor the distance between local and global prototypes to further improve the performance. To realize personalized federated learning, a hypernetwork is deployed on server to dynamically update different layers’ weights of local model. Experimental results on benchmark datasets show that our method outperforms state-of-the-art methods.
\end{abstract}
\section{Introduction}
\label{sec:intro}

With a large number of texts, images, videos, and other media data being generated, it is particularly important to conduct similarity search for multimedia data reasonably and effectively  \cite{wang2020batch,jose2022deep,liu2022deep}. In real applications, users often need to use data from one modality (e.g., text modality) to retrieve relevant data from another modality (e.g., image modality). Benefiting from high retrieval speed and low storage cost, cross-modal hashing attracts increasing attention. Cross-modal hashing (CMH) maps high-dimensional raw data to short binary hash codes by learning hash functions, while maintaining the similarity of the original samples in Hamming space. Although cross-modal hashing \cite{luo2019discrete,shen2015supervised,jiang2017deep} has achieved satisfactory performance so far, they are encountering some practical problems due to the growing concern about privacy protection. In many real-world situations, multimedia data is scattered across multiple silos and those distributed data may not be directly shared or collected due to privacy concerns and  regulations. This makes each client use only its own small amount of data for independent local training, which significantly degrades cross-modal hashing performance compared to traditional centralized training. To address the above issues, researchers try to combine federated learning  \cite{zhao2018federated,yu2020federated,liu2021quantitative,jhunjhunwala2021adaptive} with cross-modal hashing.

At present, federated cross-modal hashing is intractable due to the following challenges. 1) \textbf{Statistical heterogeneity.} The local data distribution of each client varies with its location and preferences, resulting in the data of each client being independent but not obeying the same distribution (Non-IID) \cite{mcmahan2017communication,sattler2020clustered}. 2) \textbf{Model heterogeneity.} Traditional federated learning requires consistency of models across all clients, which is unrealistic in practical and complex applications \cite{zong2021fedcmr,li2020federated}. Different clients desire different models because their own application scenarios may differ. However, different models may cause huge difficulty for the communication of parameters between clients and central server. 3) \textbf{Personalized federated learning.} Most of prior efforts let central server acquire and process parameters from all clients first and then return the same parameters \cite{huang2021personalized,tan2022towards} to all clients. The policy that each client is forced to accept the same parameters prevents each client from better adapting to its own local data, resulting in sub-optimal performance.

To address above-mentioned challenges, we propose a new federated cross-modal hashing method called Prototype-based Layered Federated Cross-Modal  Hashing (PLFedCMH for short). Specifically, on the basis of class-wise hash codes, PLFedCMH introduces class prototypes generated by modality networks to assist the learning of supervised hash functions, reducing the impact of statistical heterogeneity (non-IID) on different clients. Besides, the server only needs to aggregate local class prototypes and does not need to aggregate model parameters. There is no need to consider the parameter aggregation problem caused by model heterogeneity. Last but not least, PLFedCMH is designed with personalized federated strategy. Through the hypernetwork introduced by the server, the weights of different layers on the client are dynamically updated, which can realize personalized parameter customization for different clients. To summarize, the main contributions of this paper are as follows. 1) To consider privacy concerns, a new federated method PLFedCMH is elaborately designed for training cross-modal hashing with distributed data. 2) The proposed method could simultaneously take statistical heterogeneity, model heterogeneity, and personalized federated learning into consideration. 3) Experimental results on benchmark datasets show that our method can achieve significantly improved accurate in IID, nonIID-equal, and nonIID-unequal scenarios.

\vspace{-2mm}
\section{Proposed method}
\label{sec:format}
\vspace{-2mm}
\subsection{Problem Definition and Notations}
\label{ssec:subhead}
Following existing cross-modal hashing literature and without loss of generality, we formulate our model in the context of image-text retrieval task. In this paper, we propose a new federated learning method for cross-modal hashing, which could support hashing model training based on different silos' data without privacy and security concerns.

Assuming there are $\mathit{N}$ clients, $i$-th client possesses its own dataset $\mathit{D_i=\{(\mathbf{x}_j^{(i)},\mathbf{t}_j^{(i)},\mathbf{l}_j^{(i)})\}_{j=1}^{m_i} (1 \leq i \leq N)}$, where $\mathit{m_i}$ denotes the number of samples, $\mathit{\mathbf{x}_j}^{(i)}$ ($\mathit{\mathbf{t}_j}^{(i)}$) represents image modality (text modality) of the $\mathit{j}$-th sample on client $i$, $\mathit{\mathbf{l}_j}^{(i)}$ is the label vector. The size of all clients' datasets can be obtained by $\mathit{M=\sum_{i=1}^N m_i}$. 

As our method introduces and leverages prototypes of classes, we let $\mathbf{P}$ be the class prototype and then have both local prototypes $\mathit{\mathbf{P}_{local} : \{ \mathbf{P}_{local_1}, \mathbf{P}_{local_2}, \cdots, \mathbf{P}_{local_N} \}}$ and global prototype $\mathit{\mathbf{P}_{global}}$. Considering both image and text modalities, we have $\mathbf{P}_{local_{x_i}}$ and $\mathbf{P}_{local_{t_i}}$ on client $i$, while $\mathbf{P}_{global_{x}}$ and $\mathbf{P}_{global_{t}}$ on server. Specifically, local prototypes are output values of modality networks' last layer without using activation function. Global prototype is the average of all local prototypes, which is computed on the server.

\vspace{-2mm}
\subsection{Similarity Preserving based on Local Data}
\label{ssec:subhead}
As the core of learning to hash is to preserve the similarity, hash codes of those instances, which belong to the same class, should be relatively similar. In other words, hash codes of instances have direct relation with their labels  $\mathit{\mathbf{L} \in \{ 0, 1\} ^ {M \times c} }$, where $\mathit{c}$ is the number of classes for all data samples $\mathit{M}$.

To construct and preserve such relation, we first introduce hash codes of classes which is denoted as  $\mathit{\mathbf{Y} \in \{-1,1\}^{r \times c}}$ where $r$ is the hash code length. Then, we define the following optimization problem: $\ {min}_{\{\mathbf{B}^{(i)},\mathbf{Y}^{(i)}\}} \  \lVert r\mathbf{L}^{(i)} - {\mathbf{B}^{(i)}}^T\mathbf{Y}^{(i)} \rVert _F^2,$
where superscript $i$ denotes $i$-th client, $\mathit{\mathbf{L}^{(i)}}$ is its label matrix, $\mathit{\mathbf{B}^{(i)}}\in \{-1,1\}^{r \times m_i}$ is the hash codes, and $\mathit{\mathbf{Y}^{(i)}}\in \{-1,1\}^{r \times c}$ is the hash codes of classes on the $\mathit{i}$-th client. The above equation could force the hash codes of samples which share same labels to be more similar and thus achieve the similarity-preserving goal.

\vspace{-2mm}
\subsection{Global Information Embedding}
\label{ssec:subhead}
Sec.2.2 works only with clients' own local data. Under the influence of statistical heterogeneity (non-IID), a single client cannot well consider the overall class characteristics of the entire dataset $D$. If no remedy is taken, the local client may get stuck in its seen classes and cannot handle samples of unseen classes which are distributed on other clients.

To overcome such limitation caused by non-IID, we try to embed global information into local training. Thus, we define the following optimization function which let the local hash codes interact with global class prototypes: 
$ {min}_{\mathbf{B}^{(i)}} \  \lVert r\mathbf{L}^{(i)}_i - {\mathbf{B}^{(i)}}^T\mathbf{P}_{global} \rVert _F^2,$
where $\mathit{\mathbf{B}^{(i)}}\in \{-1,1\}^{r \times m_i}$ is the hash code matrix of samples on the $\mathit{i}$-th client, $\mathbf{P}_{global}$ is class prototypes aggregated on the server in last federated round.

In addition, we also try to keep the consistency of class prototypes between one local client and the global server. This could reduce the influence of class distribution differences on local training and improve the accuracy of local cross-modal hashing retrieval. The loss function is as follows,
\begin{equation}\label{eqn-1}
	\begin{aligned}
		&\mathit{O_1} = MSE(\mathbf{P}_{local_i}, \mathbf{P}_{global}),
	\end{aligned}
\end{equation}
where $MSE(\cdot)$ is the mean square  error and $\mathit{\mathbf{P}_{local_i}}$ is the local class prototypes generated in current federated round.

\vspace{-2mm}
\subsection{Hash Learning for Local Clients}
\label{ssec:subhead}
As most recent cross-modal hashing methods are deep ones, we could freely assume that there exist image and text modality networks. On the $i$-th client, let $\mathbf{F}^{(i)}=f(\mathbf{x}^{(i)};\pmb{\theta}_{x_i})$ denote the extracted image features, where $\pmb{\theta}_{x_i}$ represents the parameter of image modality network. For text modality, let $\mathbf{G}^{(i)}=g(\mathbf{t}^{(i)};\pmb{\theta}_{t_i})$ and $\pmb{\theta}_{t_i}$ represent the output and the parameter of text modality network of client $i$.

Then, as deep hashing could synchronously conduct hash code learning and feature extraction, based on Sec.2.2 and Sec.2.3, we could give the following equation,
\begin{equation}\label{eqn-2}
	\begin{aligned}
		\mathit{O_2} &= \alpha (\lVert r\mathbf{L}^{(i)} - {\mathbf{F}^{(i)}}^T\mathbf{Y}^{(i)} \rVert _F^2 + \lVert r\mathbf{L}^{(i)} - {\mathbf{G}^{(i)}}^T\mathbf{Y}^{(i)} \rVert _F^2)+\\
		& \beta (\lVert r\mathbf{L}^{(i)} - {\mathbf{F}^{(i)}}^T\mathbf{P}_{global_{x}} \rVert _F^2 + \lVert r\mathbf{L}^{(i)} -{\mathbf{G}^{(i)}}^T\mathbf{P}_{global_{t}} \rVert _F^2)  \\
		&+ \mu (\lVert \mathbf{B}^{(i)} - \mathbf{F}^{(i)} \rVert_F^2 + \lVert \mathbf{B}^{(i)} - \mathbf{G}^{(i)} \rVert_F^2),\\
		&s.t. \ \ \mathbf{B}^{(i)} \in \{-1,1\}^{r \times m_i}, \mathbf{Y}^{(i)} \in \{-1,1\}^{r \times c},&
	\end{aligned}
\end{equation}
where $\alpha$, $\beta$, and $\mu$ are the trade-off parameters.

\vspace{-2mm}
\subsection{Overall Loss and Optimization for Local Clients}
To make the to-be-learnt hash codes preserve both intra-client similarity and inter-client similarity, we combine Eq.\eqref{eqn-1} and Eq.\eqref{eqn-2}. Besides, as PLFedCMH is a federated method which tries to help existing CMH accommodate to distributed scenario, we should include the original loss of deep CMH $\mathit{O_{hash}}$. Thus, the overall objective function is,
\begin{equation}\label{eqn-3} 
\vspace{-2mm}
	\begin{aligned}
		&\underset{}{min} \ \ \ \  \mathit{O_1} + \eta \mathit{O_2} + \xi \mathit{O_{hash}},\\
	\end{aligned}
\end{equation}
where $\eta$ and $\xi$ are trade-off parameters.

The optimization of $\mathit{O_1} + \eta \mathit{O_2}$ could easily follow the strategy of most existing deep CMH methods, that is iteratively optimizing one variable with the others fixed. When updating network parameters $\pmb{\theta}_{x_i}$ and $\pmb{\theta}_{t_i}$, the back-propagation algorithm could be leveraged. $\mathit{\mathbf{B}^{(i)}}$ could be computed by $\mathit{\mathbf{B}^{(i)}=sign(\mathbf{F}^{(i)}+\mathbf{G}^{(i)})}$. We could use the optimization of \cite{shen2015supervised} to discretely generate $\mathit{\mathbf{Y}^{(i)}}$ bit by bit.

\begin{algorithm}[h]
	\renewcommand{\algorithmicrequire}{\textbf{Input:}}
	\renewcommand{\algorithmicensure}{\textbf{Output:}}
	\caption{PLFedCMH Algorithm}
	\label{alg1}
	\begin{algorithmic}[1]
		\REQUIRE  \{$D_1$,  $\cdots$, $D_N$ \}. Total communication rounds R. Hypernetwork learning rate $\gamma$.
		\ENSURE Trained personalized models  $ \{\bar{\pmb{\theta}}_{x_1},\cdots,\bar{\pmb{\theta}}_{x_N}\} $ and $ \{\bar{\pmb{\theta}}_{t_1},\cdots,\bar{\pmb{\theta}}_{t_N}\} $.
		\renewcommand{\algorithmicrequire}{\textbf{Server executes:}}
		\REQUIRE
\STATE  Initialization
		\FOR{each federated  round $ r \in \{ 1,\cdots,R\} $}
		\FOR{each client i $ \textbf{in parallel} $}
		\STATE $ \bar{\pmb{\theta}}_{i}^{(r+1)} = \{{\pmb{\theta}}_i^1,{\pmb{\theta}}_i^2,\cdots,{\pmb{\theta}}_i^K\} * HN_i(\mathbf{s}_i, \pmb{\zeta}_i) $
		\STATE $ \Delta\pmb{\theta}_{x_i}, \Delta\pmb{\theta}_{t_i} \leftarrow LocalUpdate(\bar{\pmb{\theta}}_{i}^{(r+1)}) $

		\STATE $ {\{{\pmb{\theta}}_{x_i}^1,{\pmb{\theta}}_{x_i}^2,\cdots,{\pmb{\theta}}_{x_i}^K\}}^{(r+1)} = {\pmb{\theta}_{x_i}}^{(r)} + \Delta\pmb{\theta}_{x_i} $
		\STATE $ {\{{\pmb{\theta}}_{t_i}^1,{\pmb{\theta}}_{t_i}^2,\cdots,{\pmb{\theta}}_{t_i}^K\}}^{(r+1)} = {\pmb{\theta}_{t_i}}^{(r)} + \Delta\pmb{\theta}_{t_i} $
		\STATE $ \mathbf{s}_i^{(r+1)} = \mathbf{s}_i^{(r)} - \gamma\nabla_{\mathbf{s}_i^{(r)}}$Eq.\eqref{eqn-3} 
		\STATE $ \mathbf{\zeta}_i^{(r+1)} = \mathbf{\zeta}_i^{(r)} - \gamma\nabla_{\mathbf{\zeta}_i^{(r)}} $Eq.\eqref{eqn-3}
		\ENDFOR		
		\STATE Update global prototypes
		\ENDFOR
		\renewcommand{\algorithmicensure}{\textbf{LocalUpdate}}
		\ENSURE $(\bar{\pmb{\theta}}_{i}^{(r+1)})$:
		\STATE Receive $(\bar{\pmb{\theta}}_{i}^{(r+1)})$ from server.
		\STATE Set $ \pmb{\theta}_{x_i} = (\bar{\pmb{\theta}}_{x_i}^{(r+1)}), \pmb{\theta}_{t_i} = (\bar{\pmb{\theta}}_{t_i}^{(r+1)}) $
		\FOR{each local epoch}
		\FOR{batch$(\mathbf{x}^{(i)}, \mathbf{t}^{(i)}, \mathbf{l}^{(i)}) \in D_i$}
		\STATE Update local prototypes. Update $ \pmb{\theta}_{x_i}, \pmb{\theta}_{t_i} $,  $\mathbf{Y}$, and $\mathbf{B}$ 
		\ENDFOR
		\ENDFOR
		\STATE \textbf{return} $ \Delta\pmb{\theta}_{x_i} = \pmb{\theta}_{x_i} - (\bar{\pmb{\theta}}_{x_i}^{(r+1)}), \Delta\pmb{\theta}_{t_i} = \pmb{\theta}_{t_i} - (\bar{\pmb{\theta}}_{t_i}^{(r+1)}) $
	\end{algorithmic}
	\end{algorithm}

\vspace{-2mm}
\subsection{Generating layered weights through the server}
\label{ssec:subhead}
For each client's local image model and text model, we set up the corresponding hypernetworks on the server side, which are composed of some fully connected layers. 

Taking the image modality of $i$-th client as an example, we have the hypernetwork $\mathit{HN_{x_i}(\mathbf{s}_{x_i},\pmb{\zeta}_{x_i})}$ \cite{ha2016hypernetworks,ma2022layer}. The input of the hypernetwork is the embedding vector $\mathit{\mathbf{s}_{x_i}}$, and $\pmb{\zeta}_{x_i}$ is the parameter of the hypernetwork. The model parameters of the image modality before updating on the server are $\mathit{\pmb{\theta}_{x_i}=\{ \pmb{\theta}_{x_i}^1, \cdots, \pmb{\theta}_{x_i}^k, \cdots,\pmb{\theta}_{x_i}^K\}}$, where $\pmb{\theta}_{x_i}^k$ are the parameters of $\mathit{k}$-th layer ($1 \leq k \leq K$). When the parameters of the client image modality are uploaded to the server, the server updates the layered parameters of the client through the image modality hypernetwork $\mathit{HN_{x_i}(\mathbf{s}_{x_i}, \pmb{\zeta}_{x_i})}$:
$\bar{\pmb{\theta}}_{x_i}= \{ \bar{\pmb{\theta}}_{x_i}^1, \cdots, \bar{\pmb{\theta}}_{x_i}^K \}= \{ \pmb{\theta}_{x_i}^1,  \cdots, \pmb{\theta}_{x_i}^K \} * HN_{x_i}(\mathbf{s}_{x_i}, \pmb{\zeta}_{x_i})$.
So the parameters $\pmb{\theta}_{x_i}$ of the $\mathit{i}$-th client are updated to $\bar{\pmb{\theta}}_{x_i}$  before the next round of local training. According to the chain rule, we can have the gradient of $\mathit{\mathbf{s}_i}$ and $\mathit{\pmb{\zeta}_i}$ from Eq. \eqref{eqn-3}.

\begin{table*}[t]
	\centering  \scriptsize  \vspace{-2mm}
	\caption{The MAP results of various methods on FashionVC and Ssense with different splits over clients. }
	\label{tab:1}
	\resizebox{\linewidth}{!}{
		\begin{tabular}{l cccccc cccccc cccccc}
			\toprule
			\multirow{3}{*}{FashionVC} & \multicolumn{6}{c}{nonIID-equal} & \multicolumn{6}{c}{nonIID-unequal} & \multicolumn{6}{c}{IID} \\
			\cmidrule(lr){2-7}\cmidrule(lr){8-13}\cmidrule(lr){14-19}
			& \multicolumn{3}{c}{Image-to-Text} & \multicolumn{3}{c}{Text-to-Image} & \multicolumn{3}{c}{Image-to-Text} & \multicolumn{3}{c}{Text-to-Image} & \multicolumn{3}{c}{Image-to-Text} & \multicolumn{3}{c}{Text-to-Image}  \\
			\cmidrule(lr){2-4}\cmidrule(lr){5-7}\cmidrule(lr){8-10}\cmidrule(lr){11-13}\cmidrule(lr){14-16}\cmidrule(lr){17-19}
			& 16bit & 32bit & 64bit & 16bit & 32bit & 64bit & 16bit & 32bit & 64bit & 16bit & 32bit & 64bit & 16bit & 32bit & 64bit & 16bit & 32bit & 64bit \\
			\midrule
			centralized &0.766 &0.762 &0.756 &0.937 &0.949 &0.946 &0.766 &0.762 &0.756 &0.937 &0.949 &0.946 &0.766 &0.762 &0.756 &0.937 &0.949 &0.946 \\
			FedAvg \cite{mcmahan2017communication} &0.398 &0.583 &0.637 &0.378 &0.583 &0.632 &0.584 &0.712 &0.741 &0.753 &0.912 &0.932 &0.544 &0.724 &0.743 &0.666 &0.903 &0.926 \\
			FedCMR \cite{zong2021fedcmr} &0.180 &0.404 &0.372 &0.159 &0.302 &0.268 &0.577 &0.620 &0.661 &0.576 &0.613 &0.707 &0.255 &0.679 &0.677 &0.229 &0.767 &0.712 \\
			FedProx \cite{li2020federated} &0.261 &0.584 &0.643 &0.228 &0.586 &0.633 &0.548 &0.710 &0.747 &0.706 &0.900 &0.930 &0.603 &0.720 &0.741 &0.770 &0.901 &0.915 \\
			FedProto \cite{tan2022fedproto} &0.297 &0.644 &0.656 &0.284 &0.659 &0.677 &0.603 &0.744 &0.760 &0.742 &0.936 &\textbf{0.947} & 0.678 &0.743 &0.761 &0.858 &0.934 &0.942 \\
			PLFedCMH &\textbf{0.614} &\textbf{0.652} &\textbf{0.657} &\textbf{0.631} &\textbf{0.671} &\textbf{0.702} &\textbf{0.710} &\textbf{0.766} &\textbf{0.763} &\textbf{0.876} &\textbf{0.941} &\textbf{0.947} &\textbf{0.731} &\textbf{0.761} &\textbf{0.769} &\textbf{0.900} &\textbf{0.947} &\textbf{0.951} \\
			\bottomrule
		\end{tabular}	}
\end{table*}

\begin{table*}[t]
	\centering \scriptsize  \vspace{-3mm}
		\label{tab:2}
	\resizebox{\linewidth}{!}{
		\begin{tabular}{l cccccc cccccc cccccc}
			\toprule
			\multirow{3}{*}{Ssense} & \multicolumn{6}{c}{nonIID-equal} & \multicolumn{6}{c}{nonIID-unequal} & \multicolumn{6}{c}{IID} \\
			\cmidrule(lr){2-7}\cmidrule(lr){8-13}\cmidrule(lr){14-19}
			& \multicolumn{3}{c}{Image-to-Text} & \multicolumn{3}{c}{Text-to-Image} & \multicolumn{3}{c}{Image-to-Text} & \multicolumn{3}{c}{Text-to-Image} & \multicolumn{3}{c}{Image-to-Text} & \multicolumn{3}{c}{Text-to-Image}  \\
			\cmidrule(lr){2-4}\cmidrule(lr){5-7}\cmidrule(lr){8-10}\cmidrule(lr){11-13}\cmidrule(lr){14-16}\cmidrule(lr){17-19}
			& 16bit & 32bit & 64bit & 16bit & 32bit & 64bit & 16bit & 32bit & 64bit & 16bit & 32bit & 64bit & 16bit & 32bit & 64bit & 16bit & 32bit & 64bit \\
			\midrule
			centralized &0.953 &0.970 &0.968 &0.973 &0.987 &0.985 &0.953 &0.970 &0.968 &0.973 &0.987 &0.985 &0.953 &0.970 &0.968 &0.973 &0.987 &0.985 \\
			FedAvg \cite{mcmahan2017communication} &0.690 &0.911 &0.929 &0.662 &0.914 &0.934 &0.865 &0.942 &0.955 &0.886 &0.971 &0.974 &0.822 &0.951 &0.962 &0.837 &0.973 &0.981 \\
			FedCMR \cite{zong2021fedcmr} &0.155 &0.343 &0.566 &0.145 &0.240 &0.379 &0.807 &0.815 &0.806 &0.760 &0.826 &0.779 &0.738 &0.821 &0.941 &0.732 &0.774 &0.941 \\
			FedProx \cite{li2020federated} &0.791 &0.908 &0.903 &0.729 &0.904 &0.888 &0.831 &0.947 &0.957 &0.855 &0.971 &0.975 &0.871 &0.948 &0.960 &0.894 &0.973 &0.980 \\
			FedProto \cite{tan2022fedproto} &0.827 &0.924 &0.917 &0.834 &0.927 &0.919 &0.936 &0.953 &\textbf{0.960} &0.973 &0.979 &0.980 &0.872 &0.956 &0.959 &0.895 &0.980 &0.983 \\
			PLFedCMH &\textbf{0.872} &\textbf{0.932} &\textbf{0.937} &\textbf{0.873} &\textbf{0.946} &\textbf{0.953} &\textbf{0.947} &\textbf{0.956} &0.958 &\textbf{0.977} &\textbf{0.981} &\textbf{0.982} &\textbf{0.957} &\textbf{0.961} &\textbf{0.963} &\textbf{0.981} &\textbf{0.983} &\textbf{0.984} \\
			\bottomrule
		\end{tabular}	}
\end{table*}

\vspace{-2mm}
\subsection{Overall  Algorithm and Framework of PLFedCMH}
\label{ssec:subhead}
Algorithm \ref{alg1} shows a federated round of the proposed PLFedCMH. 1) On the server side, two hypernetworks corresponding to different modality networks are used to generate the layered weights of all clients. 2) The server transmits the updated client layered weights and the aggregated abstract global class prototypes to clients. 3) The client updates the personalized model parameter values for modality networks after receiving the layered weights. 4) On the local client, features of the private samples are extracted through the image and text modality networks to obtain the rich semantic information of samples and the class prototypes of the local client. 5) After local training, the local model parameter updates and the local class prototypes for both modalities are uploaded to the server. 6) The server aggregates the local prototypes to obtain global prototypes. And the hypernetworks calculate the layered weights through the gradient change of the model.

\vspace{-2mm}
\section{Experiment}
\label{sec:pagestyle}
\subsection{Experiment settings}
\label{ssec:subhead}
\textbf{Datasets.} Following existing literature \cite{sun2019supervised,zhan2020supervised}, two benchmark datasets are chosen for evaluation, i.e., FashionVC \cite{song2018neural} and Ssense \cite{sun2019supervised}. FashionVC is from online fashion community Polyvore. After removing categories with less than 25 samples, FashionVC contains 19,862 image-text pairs with hierarchical labels. Ssense is also from the fashion field, which contains 15,696 hierarchically labeled image-text pairs after removing categories with less than 70 samples. In this paper, only the most fine-grained part of the hierarchical labels is employed for evaluation. 

Following setting in \cite{tan2022fedproto}, three ways to partition the data set are used, i.e.,  nonIID-equal, nonIID-unequal, and IID. NonIID-equal and nonIID-unequal are cases of statistical heterogeneity. In the nonIID-equal case, each client has a different data distribution, and the classes may overlap or not overlap at all between different clients. However, the number of categories is the same for each client, and the number of samples in each category is also the same. Given the different number of samples in each class in the dataset, we needed to accommodate the smaller classes in order to achieve nonIID-equal, so the total number of samples from all clients we used is 18\% to 20\% of the entire dataset. In the nonIID-unequal case, the dataset is 100\% used by the clients, and the number of samples per class is completely different. In the IID case, data is shuffled and then evenly divided among the clients, which have the same data distribution.

\textbf{Experiment details.} As this paper focuses on federated learning for CMH, we utilized existing SOTA SHDCH \cite{zhan2020supervised} for ours and all federated baselines. The hypernetworks for both modalities contain four fully-connected layers with ReLU activation function. The hyper-parameters are set as follows:  $\alpha=0.5$, $\beta=0.5$, $\mu=10$, $\eta=10^{-5}$, and $\xi=1$. The learning rate of modality networks is $0.0001$ and the learning rate of hypernetworks is $0.001$.

\textbf{Evaluation metrics.} Two cross-modal retrieval tasks are conducted. ``Image-to-Text'' task utilizes an image as a query to retrieve right texts, and ``Text-to-Image'' task is to retrieve desired images with a text query. We adopted the widely used Mean Average Precision (MAP) to evaluate the performance, where higher values indicate better performance.

\begin{table}[t]
	\centering \scriptsize  \vspace{-2mm}
	\caption{The MAP results on Ssense.}
	\label{tab:3}{
		\begin{tabular}{lcccccc}
			\toprule
			\multirow{2}{*}{Method} & \multicolumn{3}{c}{Image-to-Text} & \multicolumn{3}{c}{Text-to-Image}  \\
			\cmidrule(lr){2-4}\cmidrule(lr){5-7}
			& 16bit & 32bit & 64bit & 16bit & 32bit & 64bit \\
			\midrule
			PLFedCMH &\textbf{0.947} &\textbf{0.956} &\textbf{0.958} &\textbf{0.977} &\textbf{0.981} &\textbf{0.982} \\
			PFedCMH &0.938 &0.952 &0.957 &0.963 &0.978 &0.980 \\
			LFedCMH &0.938 &0.940 &0.951 &0.965 &0.966 &0.976 \\
			\bottomrule
		\end{tabular}	}
\end{table}

\vspace{-2mm}
\subsection{Comparison with Baselines}
\label{ssec:subhead}
The results of MAP values on FashionVC and Ssense datasets under nonIID and IID settings are presented in Table \ref{tab:1}. We compared the MAP values of PLFedCMH with several SOTA baselines, including FedAvg \cite{mcmahan2017communication}, FedCMR \cite{zong2021fedcmr}, FedProx \cite{li2020federated} and FedProto \cite{tan2022fedproto}. Furthermore, results of ``centralized'' are also provided, which denotes the result of accumulating all data learned on a single server and is the upper bound of the federated learning algorithm.

As found from Table \ref{tab:1}, MAP values of all baselines with 16-bit hash codes under nonIID-equal setting can only reach less than 40\% of centralized's performance due to poor characterization capability of short-bit hash codes, while our method reaches a MAP value of 61.4\%. One possible reason is that our method introduces class prototypes to assist in the learning of hash codes, which reduces the effect of statistical heterogeneity on different clients. Besides, our method achieves the best results in most cases, which implies the effectiveness of using prototypes to learn the similarity between instances and classes on the server.

\vspace{-2mm}
\label{ssec:subhead}
\subsection{Ablation experiments}
\label{ssec:subhead}

To fully validate the performance of PLFedCMH, two variants are designed. The first variant removes layered update weights on the server to different clients, which is named PFedCMH. The other variant is termed LFedCMH, which excludes the prototypes. Comparison results  with nonIID-unequal split are listed in Table \ref{tab:3}. From those tables, we can find our PLFedCMH could perform better than two designed variants. Such phenomena reveal that both updating layered parameters and using prototypes to learn the similarity between instances and classes on the server are effective.

\vspace{-2mm}
\label{ssec:subhead}
\section{Conclusion}
\label{sec:prior}
In this paper, we propose a novel federated learning method PLFedCMH for cross-modal hashing with distributed data. We introduce class prototypes generated by modal networks to assist the hash learning, reducing the impact of statistical heterogeneity (non-IID) on different clients. At the same time, distance between local and global prototypes is considered to improve the performance. The server dynamically updates the weights of different layers of the client, which can realize personalized parameter customization for different clients. On the other hand, the server only needs to aggregate local category prototypes without aggregating model parameters, reducing the impact of model heterogeneity. Experimental results show that the proposed method achieves the best performance on benchmark datasets.

\vfill\pagebreak

\clearpage
\bibliographystyle{IEEEbib}
\bibliography{Template}

\end{document}